# Overcoming Misleads In Logic Programs by Redefining Negation

M. A. El-Dosuky [a], T. T. Hamza [a], M. Z. Rashad [a] and A. H. Naguib [b]

[a] Faculty of Computers and Information sciences, Mansoura University, Egypt
E-mail: mouh_sal_010@mans.edu.eg

[b] Faculty of Sciences, Mansoura University, Egypt

Negation as failure and incomplete information in logic programs have been studied by many researchers In order to explains HOW a negated conclusion was reached, we introduce and proof a different way for negating facts to overcoming misleads in logic programs. Negating facts can be achieved by asking the user for constants that do not appear elsewhere in the knowledge base.

## 1. Introduction

Negation as failure and incomplete information in logic programs have been studied by many researchers, mainly because of their role in the foundations of declarative reading of logic programming. This paper gives a review of some of the definitions of the concepts related to of the declarative reading of logic programming. Then, the paper provides a framework to overcome misleads and to solve a misleading case study.

The paper begins with reviewing the relevant work of contributions to logic programming emphasizing many concepts such as negation as failure, closed world assumption, incomplete information, and their consequences (Section 2). Then we comment on the standard definitions of the relevant logic programming concepts such as: compound terms, substitution, common instance, facts, rules, reduction, variables quantification, unifier, Most General Unifier (MGU), computation, and structured data (Section 3).
Then we briefly discuss the semantics of logic programming. A logic program can have many semantics according the point of view. The common semantics are operational, denotational, and declarative (Section 4). Then we present our framework for overcoming misleads in logic programs using negation as invalid (Section 5). Then we investigate the features of the presented framework in the next section.

## 2. Related work

Different research communities have investigated various aspects of the declarative readings of logic programming.
Logic programming languages are a great tool for knowledge representation. The main non-monotonic feature is **Negation As Failure** [Clark1978] . It enables to express knowledge not readily expressible in classical logic. However, these languages cannot directly represent incomplete knowledge about the world.
A consistent classical theory distinguishes between three types of sentences : provable, refutable, and undecidable.  However, a logic program answers either yes or no, with no counterpart for undecidable sentences which represent the incompleteness of information. The reason for this is due to the declarative semantics of logic programming automatically applies the **closed world assumption** [Reiter1978]. It states that each ground atom that does not follow from the facts in the program is assumed to be false. The causal logic [Bochman 2004] is shown to make any causal logic program satisfies the **Negation As Default** principle (alias Closed World Assumption).
Many attempts are made to allow for the correct representation of **incomplete information**. The first is probably due to Jack Minker [Minker1982] . Another  attempt is to expand the language of logic programs with classical negation as failure by modal operators K and M defined with details in [Gelfond1994]. For the set of rules T and formula F, KF stands for "F is known to be true by a reasoner with a set of premises T" and MF means "F may be believed to be true". Incomplete information can be captured by communicating other intelligent parities [Rosati2003]. This framework is well-suited for representing situations in which an agent cooperates in a team, and each agent is able to communicate his knowledge to other agents in the team.
Logic programming — and negation as failure — has many declarative readings. One view is that logic programming is a logic for default reasoning. In this view, negation as failure is a modal operator. In an alternative view, a logic program is interpreted as a definition. In this view, negation as failure is the classical objective negation [Denecker 2000].
There is an **epistemological ambiguity** that arises in the context of logic programming [Denecker 2004]. There are several logic programming formalisms, each has  a pair of a formal syntax and a formal semantics. [Denecker 2004] falsifies our assumptions that that each such pair has a unique declarative reading and for a program in several formal logics  shall its declarative reading is the same in each of them.

## 3. Preliminaries

**Definition: Facts**
A **relation** between objects is called a predicate . A relation is summarized in a simple phrase called a fact that consists of the relation name followed by the object or objects (enclosed in parentheses) .The facts ends with a period (.)*.* Facts can also express properties of objects as well as relations, depending on arity.
Examples :

      A relation:        Jack likes Sarah        likes( jack , sarah)
      A property:        Kermit is green        green ( kermit)

**Definition: Rules**
Rules enable you to infer facts from other facts. A rule has two parts *Head* and *Body*

      Head ← <subgoal$_1$>,<subgoal$_2$> , … , <subgoal$_N$>

The body of the rule consists one or more sub-goals .

**Definition: Compound terms**
Compound terms have the form: $f(t_1, t_2, ...., t_n)$. where f is called functor, $t_i$ is the arguments, and n is the arity. A **functor** looks like fact, but it is not a fact . It's just a data object, which you can handle in much the same way as a symbol or number. A functor does not stand for some computation to be performed. It's just a name that identifies a kind of compound data object and holds its arguments together.
Example: successor(zero)

**Definition: Structured Data**
Structured data are constructed by grouping similar arguments of a fact and giving a name to that group. Facts are the counterpart of tables, while structured data corresponds to records with aggregate fields.
Example:
    A fact about a lecture course on logic given on Monday from 9 to 11 by prof. Jack in the Victory building, room A, can be represented as
        course (logic, monday, 9, 11, jack, victory, a).
    Using *structured data*, we can define a four-argument version of this fact as
        course (logic, time(monday, 9, 11),lecturer(jack), location(victory, a)).

    In a logic language, commonly Prolog, you can do this by declaring a **domain** containing the compound object date. For example, declaring the time:
        **time_compound = time (symbol, integer, integer)**

    Using the concept of **primary key,** we can rewrite this fact in terms of *binary* relations, each takes a specific information and the course name as a primary key.
        day (logic, monday).
        start_time(logic, 9).
        finish_time(logic, 11).
        lecturer(logic, jack).
        building(logic, victory).
        room(logic, a).
It is possible to define a set of retrieval rules to check a specific piece of information within the fact.
        **lecturer**(Course, Lecturer) ← course (Course, Time, Lecturer, Location).
        **duration** (Course, Length) ←
            course (Course, time(Day, Start, Finish), Lecturer, Location),
            plus(Start, Length, Finish).
        **teaches**(Lecturer, Day) ←
            course (Course, time(Day, Start, Finish), Lecturer, Location).

**Definition: Substitution**
Substitution is a finite set of pairs of the form $X_i = t_i$. Where: $X_i$ is a variable and $t_i$ is a term. Note that $X_i ≠ X_j$ for every i ≠ j and $X_i$ does not occur in $t_j$.
Example: {X = jack}

**Definition: common instance**
C is a common instance of A and B if it is an instance of A and an instance of B. In other word, if there are substitutions $\theta_1$ and $\theta_2$ such that C=A$\theta_1$ is syntactically identical to B$\theta_2$
Example:      A = plus(0, 3, Y)            $\theta_1$ = {Y = 3}
            B = plus(0, X, X)           $\theta_2$ = {X = 3}
                    C = plus(0, 3, 3)

**Definition: Reduction**
A reduction of goal G by a program P is the replacement of G by the body of an instance of a clause in P, whose head is identical to the chosen goal. Example: the goal grandparent(abraham, X)?
Can be reduced to          parent (abraham, Y), parent(Y, X)
with respect to the clause :       grandparent(Z, X) ← parent (Z, Y), parent(Y, X).

**Definition: Variables quantification**
Variables in queries are *existentially* quantified, while in facts are *universally* quantified. A query $p(T_1,T_2,…,T_n)?$ , which contain the variables $X_1, X_2, ..., X_k$ reads:
"Are there $X_1, X_2, ..., X_k$ such that $p(T_1,T_2,..., T_n)?$ " usually, existential quantification is usually omitted.
Example: the query father (abraham, X)? reads :
               "Does there exist an X such that abraham is the father of X ? "
A fact $p(T_1,T_2,…,T_n)$ reads : "For all $X_1, X_2, ..., X_k$ ,where $X_i$ are the variables occurring in the fact $p(T_1,T_2,..., T_n)$ , is true ". Logically, from a universally quantified fact one can deduce any instance of it.
Example: From likes(X, apple) , deduce likes (abraham, apple)

**Definition: Most General Unifier (MGU)**
A unifier of two terms is a substitution making the terms identical. If two terms have a unifier, we say they unify. MGU of two terms is a unifier such that the associated common instance is most general. If two terms unify, all MGUs are equivalent.

**Definition: Computation**
a computation of a goal $Q= Q_0$ by a program P is a sequence of triples $(Q_i, G_i, C_i)$ .
where : $Q_i$ is a conjunctive goal, $G_i$ is a goal occurring in $Q_i$, and $C_i$ is a clause $A←B_1,…, B_k$ in P, that contains new variables not occurring in $Q_j$,  $0≤j≤i$.

For a detailed introductory discussion please refer to  ([Sterling& Shapiro 1994] and [Bochman1998]).

## 4.Semantics of Logic Programs
A logic program can have many semantics according the point of view. The common semantics are operational, denotational, and declarative. The operational semantics of logic programs are considered a way of describing the meaning of the programs procedurally. It is a set of ground goals that are instances of queries solved by a logic program P using abstract interpretation as shown in section 4.1. The declarative semantics of logic programs are derived from the term model, referred to as the Herbrand base. The declarative semantics is discussed in section 4.2.
The denotational semantics of logic programs are defined in terms of a function which assigns meaning to the program. The function is over the domain computed by the program. Meaning is defined as the least fixpoint of the function, if it exists. The denotational semantics is out of concern of this paper, since Kowalski's famous slogan "Algorithms= Logic + Control" [Kowalski1979] implies that declarative logic statements can be interpreted as procedural computer instructions. We presume that declarative and operational semantics are sufficient to reflect the algorithm of a logic program.

### 4.1. Interpreting logic programs
In the next, we show the algorithm of an abstract interpreter for logic programs. It is used to construct the operational semantics for a logic program. Note that this algorithm depends on **unify** function which is mentioned in line 5. This function is defined by the Unification algorithm.

**Abstract interpreter for logic programs**
**Input:** a goal G and a program P
**Output:** an instance of G that is a logical consequence of P, or no otherwise.
**Algorithm:**
   1        initialize the resolvent to G.
   2        **while** the resolvent is not empty **do**
   3            choose a goal A from the resolvent
   4            choose a (renamed) clause A' ← B1, …, Bn from P
   5                such that A and A' **unify** with MGU Θ.
   6            (if no such goal and clause exist, exit the while loop)
   7            replace A by B1, …, Bn in the resolvent
   8            apply Θ to the resolvent and to G
   9        if the resolvent is empty, then output G, else output no.

**Unification algorithm**
**Input:** two terms T1 and T2 to be unified
**Output:** Θ, the MGU of T1 and T2, or failure
**Algorithm:**
```
1        Initialize the substitution Θ to be empty,
2        Initialize the stack to contain the equation T1 = T2,
3        Initialize failure to false.
4        while stack not empty and no failure do
5                pop X = Y from the stack
6                case
7                        X is a variable that does not occur in Y:
8                                substitute Y for X in the stack and in Θ
9                                add X = Y to Θ
10                       Y is a variable that does not occur in X:
11                               substitute X for Y in the stack and in Θ
12                               add Y = X to Θ
13                       X and Y are identical constants or variables:
14                               continue
15                       X is f(X₁, …, Xₙ) and Y is f(Y₁, …, Yₙ)
16                       for some functor f and n > 0:
17                               push Xᵢ = Yᵢ, i = 1 … n, on the stack
18                       otherwise:
19                               failure is true
20       if failure, then output failure else output Θ
```

Tracing a logic program is a good way for capturing its declarative meaning. A trace of a computation of logic program $(Q_i, G_i, C_i)$ is a sequence of pairs $(G_i, Θ_i')$, where $Θ_i'$ is the subset of mgu $Θ_i$ computed at the ith reduction, restricted to variables in $G_i$. The meaning of a logic program is a set of ground deducible from the program. <u>Example</u>: tracing the appending of two lists.

    **append([ ], Ys, Ys).**
    **append([X|Xs],Ys,[X|Zs])← append(Xs,Ys,Zs).**
    **Trace:**

| | |
|---|---|
| append ([a, b], [c, d], Ls) | Ls = [a\|Zs] |
|   append([b], [c, d], Zs) | Zs = [b\|Zs1] |
|     append([], [c, d], Zs1) | Zs1 = [c, d] |
|       **true** | |
|       output: Ls = [a, b, c, d] | |

**Implementation of Negation as failure**
Classical negation plays a special rule in logic programming[Alferes**1998**]. Negation as failure is implemented as *not* predicate. It is a meta-logical predicate, which means that it takes predicates as its argument. To implement *not* predicate, *cut* and *fail* predicates are used as follows :
  not(X) :- X, !, fail.
  not(X).
Fail always fails when the search reaches it. It corresponds to any impossible condition as 2=3. Cut, denoted !, always succeeds when the search reaches it. Cut is procedural and cannot be understood in non-procedural terms. Cuts can be useful for reducing the search space of a program [Colmerauer& Roussel1996].

**Dynamic Database**
The bottleneck of any system written in Prolog is the knowledge base (dynamic database). An internal database is composed of facts that you can add directly into and remove from program at run time. There are three predicates to add a single fact at runtime:
- **asserta** asserts a new fact before the existing facts for the predicate,
- **assertz** asserts a new fact after the existing facts for that predicate
- **assert** behaves like assertz.

To remove facts from the databases. retract can be used with the form: retract(<the fact> )
Retract will remove the first fact in the database that matches <the fact>, instantiating any free variables in <the fact>.
For a detailed review for interpreting logic programs, you can refer to ([Sterling& Shapiro 1994] , [Colmerauer1985] , and [Krzysztof1996]).

## 4.2. Declarative reading of logic programs

One feature of the declarative reading of logic programs that makes it very useful is the justification and explanation of HOW and WHY this semantics characterizes this declarative reading [Denecker 2000]. Explanation is an important facility provided by expert systems [Merritt1989]. The system must be able to explain HOW it arrived a conclusion and WHY it is performing some computation. To answer how a conclusion was reached, work back through the inference chain. To answer why a computation is being performed, the system must state its current goal. Figure 1 applies this concept on a simple rule. Decision 1 was made true because Facts 1, 2, 3 are true. The system may ask the user if fact 3 is true, because it is trying to determine if decision 1 should be made.

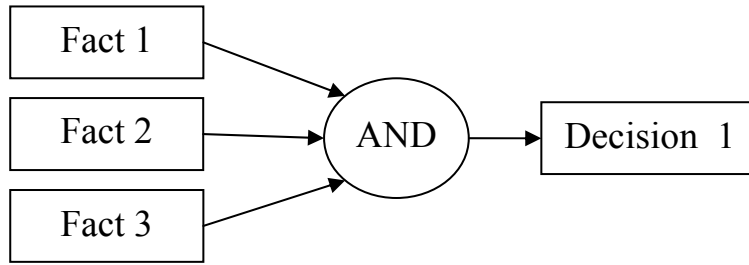

Figure 1: HOW and WHY Explanation of a simple rule

**Definition: Interpretation**
Let *P* be a logic program. Let *U(P)* be Herbrand Universe of *P* and *B(P)* be Herbrand Base of *P*. *U(P)* is the set of all ground terms that can be formed from the constants and function symbols appearing in *P*. *B(P)* is the set of all ground goals that can be formed from predicates in *P* and the terms in the Herbrand universe. An interpretation of a logic program is a subset of the Herband base.

For a review for the famous Herbrand theorem and new sequent forms of Herbrand theorem and their applications, you can refer to [Lyaletski2006].

**Definition: Model**
Let *P* be a logic program. Let *I* be an interpretation. *I* is a model for *P* if for each ground instance of a clause $A \leftarrow B_1, \ldots, B_n$ in *P*   *A* is in *I* if $B_1, \ldots, B_n$ are in *I*.

**Definition: Minimal Model**
Let *P* be a logic program. Let *M(P)* be minimal model of *P*. *M(P)* is the intersection of all models. The minimal model is the declarative meaning of a logic program

**Definition: Mapping**
Let *P* be a logic program. Let $T_P$ be natural mapping from interpretations to interpretations.
$T_P(I) = \{ A \text{ in } B(P) : A \leftarrow B1, \ldots, Bn, n \geq 0,$ is a ground instance of a clause in P, and $B1, \ldots, Bn$ are in $I\}$.

**Definition: Intended meaning**
Let *P* be a logic program. Let *M* be the intended meaning of *P*. *M* is a set of ground goals intended by programmer for the program to compute.

**Definition: Correctness**
Let *P* be a logic program. Let *M* be the intended meaning of *P*. *P* is correct with respect to *M* if *M(P)* is contained in *M*.

**Definition: Completeness**
Let *P* be a logic program. Let *M* be the intended meaning of *P*. *P* is complete with respect to *M* if *M* is contained in *M(P)*.

For a comprehensive review for declarative reading of logic programs, you can refer to ([Sterling& Shapiro 1994] , [Lyaletski2006] , and [Denecker 2000]).

## 5. Overcoming Misleads In Logic Programs by Redefining Negation
### 5.1 Motivation
Sometimes it is very hard to explain HOW a negated conclusion was reached. Consider the following verbal problem [Dow&Mayer 2004]:
> Marsha and Marjorie were born on the same day of the same month of the same year to the same mother and the same father yet they are not twins. How is that possible?

To represent that Marsha and Marjorie were born on the same day of the same month of the same year to the same mother and the same father we can use the *person* predicate as follows:
  person(marsha, father1, mother1, month1, year1),
  person(marjorie, father1, mother1, month1, year1).

To represent that twins were born on the same day of the same month of the same year to the same mother and the same father we can use the *twin* predicate as follows
  twin (A,B) ←
    person(A, Father, Mother, Month, Year),
    person(B, Father, Mother, Month, Year).
By tracing the goal of not being twins as the given example in section 4.1
  goal
    not(twin (marsha, marjorie))
we reach a conclusion of No. There is no way to explain HOW they are not twins, because of the contradicting definition of *twin* predicate. We need another representation that can provide an explanation for not being twins.

**5.2 Methodology**
In this section we provide and prove the key properties of our framework to explain HOW a negated conclusion can be reached. But first, let us review some important inference rules.

**Definition: Skolemization**
*Skolemization* is a way of eliminating existential quantifiers. Variables bound by existential quantifiers not inside the scope of universal quantifiers can be replaced by constants.
Example:
$\exists x [x<5]$ can be changed to $c<5$, with $c$ a suitable constant according to existential instantiation.
For a review for the famous Skolemization and you can refer to ([Bürckert, et al 1996] and [Skolem1970]).

**Definition: Existential Instantiation**
For any sentence $a$, variable $v$, and constant symbol $k$ that does not appear elsewhere in the knowledge base:

$$\frac{\exists v \ \alpha}{\text{SUBST}(\{v/k\}, \alpha)}$$

For example, from $\exists x\ Kill(x, Victim)$, we can infer *Kill(Murderer, Victim)*, as long as *Murderer* does not appear elsewhere in the knowledge base [Russel&Norvig 2003].

**Theorem** Negating facts can be achieved by asking the user for constants that do not appear elsewhere in the knowledge base
**Proof**
As noted in section 3, variables in facts are *universally* quantified, so a fact $p(x_1,\ldots,x_n, t_1,\ldots,t_m)$ actually means :
$$\forall x_1 \ldots \forall x_n\ p(x_1,\ldots,x_n, t_1,\ldots,t_m)$$
where $x_1..x_n$ are variables and $t_1 .. t_m$ are terms.
Negating the fact using De Morgan Laws
$$\neg \forall x_1 \ldots \forall x_n\ p(x_1,\ldots,x_n, t_1,\ldots,t_m) \equiv \exists x_1 \ldots \exists x_n\ \neg\ p(x_1,\ldots,x_n, t_1,\ldots,t_m)$$

Applying Skolemization to eliminate existentially quantified variables with constants.
$$\exists x_1 \ldots \exists x_n\ \neg\ p(x_1,\ldots,x_n, t_1,\ldots,t_m) \equiv \neg\ p(c_1,\ldots,c_n, t_1,\ldots,t_m)$$

By introducing another predicate $s^{(n+1)}$ with the form
$$s(p, c_1,\ldots,c_n, t_1,\ldots,t_m) \equiv p(c_1,\ldots,c_n, t_1,\ldots,t_m)$$

So $\exists x_1 \ldots \exists x_n\ \neg\ p(x_1,\ldots,x_n, t_1,\ldots,t_m) \equiv \neg\ p(c_1,\ldots,c_n, t_1,\ldots,t_m) \equiv s(\neg p, c_1,\ldots,c_n, t_1,\ldots,t_m)$

To ensure that constants $c_1,\ldots,c_n$ do not appear elsewhere in the knowledge base, we rely on asking the user for them and remembering the answer as described in the next section.

**5.3 Asking the user and Remembering the answer**
The **ask** predicate [Merritt1989] will have to determine from the user whether or not a given attribute-value pair is true for a specific person. A new predicate, **known** is used to remember the user's answers to questions. It is not specified directly in the program, but rather is dynamically **assert**ed whenever **ask** gets new information from the user. Every time **ask** is called it first checks to see if the answer is already **known** to be yes or no. If it is not already **known**, then **ask** will **assert** it after it gets a response from the user. The arguments to **known** are: yes/no, attribute, person, and value.

Our new version of **ask** looks like:
    ask(A, P, V) ←
        known(yes, A, P, V), !. % succeed if true and stop looking
    ask(A, P, V) ←
        known(_,A, P, V), !, fail.   % fail if false
    ask(A, P, V) ←
        write(A, " of person ", P,  " is ", V, " ? "),
        readln(Y), % get the answer
        asserta(known(Y, A, P, V)), % remember it
        Y = yes. % succeed or fail

**5.4 Case study**
Consider the verbal problem introduced in section 5.2 again. Using the concept of structured data we can define missing information to be anything that relate to the existing data and do not occur in the knowledge base. For example: consider "family" which can be derived from "father" and "mother". Also consider "day" which can be derived from "month" and "year ". Now, any two persons who share father and mother can have different families or different birth date, so they are not twins.

Another way is to add one or more new objects that do not occur in the knowledge base. Thus, any two persons who share father, mother and birth date can have another person with the same  father, mother and birth date, so they are triplets not twins.

The full corresponding PROLOG listing is shown below:
    domains
        name, father, mother, month, year = symbol
        missing =country(symbol, symbol); family(symbol, symbol); day(symbol, symbol)
    predicates
        person(name, father, mother, month, year);
        person(name, father, mother, month, year, missing)
        state (symbol, name, name)
    clauses
        person(marsha, father1, mother1, month1, year1).
        person(marjorie, father1, mother1, month1,year1).

        state (twin, A,B) ←
            person(A, Father, Mother, Month, Year),
            person(B, Father, Mother, Month, Year).
        state (not_twin, A,B) ←
            person(A, Father, Mother, Month, Year, country(X, A)),
            person(B, Father, Mother, Month, Year, country(Y, B)).
        state (not_twin, A,B) ←
            person(A, Father, Mother, Month, Year, family(X, A)),
            person(B, Father, Mother, Month, Year, family (Y, B)).
        state (not_twin, A,B) ←
            person(A, Father, Mother, Month, Year, day(X, A)),
            person(B, Father, Mother, Month, Year, day (Y, B)).
        state (not_twin, A,B) ←
            person(A, Father, Mother, Month, Year),
            person(B, Father, Mother, Month, Year),
            person(C, Father, Mother, Month, Year).

Any missing data can be acquired using the ask predicate as follows:
    country (X, P) ← ask(country, P, X).
    day (X,P) ← ask(day, P, X).

    person(P, Father, Mother, Month, Year) ←
        ask if there is a person P of father  Father, mother Mother, month Month, and year Year

A trace of the goal    state (not_twin, marsha, marjorie)?    shows how Marsha and Marjorie can share the same father, mother and birth date but still not be twins as they have they have different families or different countries.  Another way of explanation is that although they share father, mother and birth date, there is another person with the same  father, mother and birth date, so they are triplets not twins.

## 6. Conclusion and Future work

In order to explains HOW a negated conclusion was reached, we introduce negating facts in a different way. Negating facts can be achieved by asking the user for constants that do not appear elsewhere in the knowledge base.

In future work, we aim to investigate efficient implementation in detail. We aim to introduce a predicate named *not* to automatically perform the labor required.

Another promising direction for future research is to incorporate the negating methodology proposed in this paper into one of the several extensions to PROLOG. The resulting, more flexible support for representing negation may constitute an important step towards built-in support for overcoming misleads in logic programs and form a qualitative leap for introducing creativity to machines.

## References

[Alferes**1998**]    Alferes J. J., Pereira L. M., Przymusinski T. C., "'Classical' Negation in Nonmonotonic Reasoning and Logic Programming", J. Automated Reasoning, (20):107-142, 1998.

 [Bochman1998]    Bochman A., "A logical foundation for logic programming, I and II, Journal of Logic Programming*,* pages 151–194. 1998

[Bochman2004]    Bochman A.," A Causal Logic of Logic Programming", Proceedings of the Ninth International Conference (KR2004), pages 427-437,  AAAI Press, 2004

[Bürckert, et al 1996]    Bürckert H.J., Hollunder B.and Laux A. ,"On Skolemization in constrained logics , Annals of Mathematics and Artificial Intelligence , (18), pages 95-131, 1996

[Clark1978]    Clark K., "Negation as failure", In Herve Gallaire and Jack Minker, editors, Logic and Data Bases, pages 293-322, PlenumPress,  New York, 1978

[Colmerauer1985]    Colmerauer, A., "Prolog in 10 Figures", Communications of the ACM, pp. 1296-1310,1985

[Colmerauer& Roussel1996]    Colmerauer A. and Roussel P. "The birth of Prolog", dans *History of Programming Languages*, edited by Thomas J. Bergin and Richard G. Gibson, ACM Press/Addison-Wesley, 1996.

[Denecker 2000]    Denecker M., "A note on the Declarative reading of Logic Programming", In Proc of AAAI-2000, 2000

[Denecker 2004]    Denecker M., " What's in a Model? Epistemological Analysis of Logic Programming", In Proc of AAAI-2004, 2004

[Dow&Mayer 2004]    Dow, G.T. & Mayer, R.E., "Teaching students to solve insight problems.  Evidence for domain specificity in training", *Creativity Research Journal, (16), pages* 389-402,2004

[Gelfond1994]    Gelfond M., "Logic Programming and Reasoning with Incomplete Information", Annals of Mathematics and Artificial Intelligence, 12, pages 89-116,1994

[Kowalski1979]    Kowalski, R.A., "Algorithm=logic+control", Communications of the ACM, pp. 424-436, 1979

[Krzysztof1996]    Krzysztof R., "From Logic Programming to Prolog", Prentice Hall; 1st Ed, 1996

[Lyaletski2006]    Lyaletski A., "Sequent forms of Herbrand theorem and their applications" , Annals of Mathematics and Artificial Intelligence , (46) pages 191-230, 2006

[Merritt1989]    Merritt D, "Building Expert Systems in Prolog", Springer-Verlag, 1989

[Minker1982]    Minker J., "On indenite data bases and the closed world assumption", In Proc of CADE-82,  pages 292-308, 1982

[Reiter1978]     Reiter R., "On closed world data bases", In Herve Gallaire and Jack Minker, editors, Logic and Data Bases,  pages 119-140, Plenum Press, New York, 1978

[Rosati2003]    Rosati R., "Minimal belief and negation as failure in multi-agent systems". Annals of Mathematics and Artificial Intelligence, vol. 37, pages 5-32, 2003

[Russel&Norvig 2003]    Russel S. and Norvig P., "Artificial Intelligence A Modern Approach" 2[nd] Ed, page 273, Pearson Education, 2003

[Skolem1970]    Skolem, T., "Selected Works in Logic", edited by Fenstad J. E., Universitetsforlaget, 1970.

[Sterling& Shapiro 1994]    Sterling L. and Shapiro E., "The Art of Prolog: Advanced Programming Techniques", 2nd Ed., MIT Press, 1994